\documentclass[conference]{IEEEtran}

\IEEEoverridecommandlockouts
\usepackage{cite}
\usepackage{amsmath,amssymb,amsfonts}
\usepackage[capitalize]{cleveref}
\usepackage{algorithmic}
\usepackage{graphicx}
\usepackage{textcomp}
\usepackage{xcolor}
\def\BibTeX{{\rm B\kern-.05em{\sc i\kern-.025em b}\kern-.08em
    T\kern-.1667em\lower.7ex\hbox{E}\kern-.125emX}}
\renewcommand{\baselinestretch}{0.99}

\usepackage{siunitx}
\DeclareSIUnit{\nothing}{\relax}

\begin{document}

\title{Tiny-PULP-Dronets: Squeezing Neural Networks for Faster and Lighter Inference on Multi-Tasking Autonomous Nano-Drones}


\author{\IEEEauthorblockN{
Lorenzo Lamberti\IEEEauthorrefmark{1}, 
Vlad Niculescu\IEEEauthorrefmark{2},
Micha\l{} Barci\'s\IEEEauthorrefmark{3},
Lorenzo Bellone\IEEEauthorrefmark{3},
Enrico Natalizio\IEEEauthorrefmark{3},
Luca Benini\IEEEauthorrefmark{1}\IEEEauthorrefmark{2},\\
Daniele Palossi\IEEEauthorrefmark{2}\IEEEauthorrefmark{4}
}

\IEEEauthorblockA{\IEEEauthorrefmark{1} Department of Electrical, Electronic and Information Engineering - University of Bologna, Italy}
\IEEEauthorblockA{\IEEEauthorrefmark{2} Integrated Systems Laboratory - ETH Z\"urich, Switzerland}
\IEEEauthorblockA{\IEEEauthorrefmark{3} Autonomous Robotics Research Center - Technology Innovation Institute, UAE}
\IEEEauthorblockA{\IEEEauthorrefmark{4} Dalle Molle Institute for Artificial Intelligence - USI and SUPSI, Switzerland}
lorenzo.lamberti@unibo.it,
vladn@iis.ee.ethz.ch,
michal.barcis@tii.ae,
lorenzo.bellone@tii.ae,\\
enrico.natalizio@tii.ae, 
luca.benini@unibo.it,
daniele.palossi@idsia.ch
}

\maketitle

\begin{abstract}
Pocket-sized autonomous nano-drones can revolutionize many robotic use cases, such as visual inspection in narrow, constrained spaces and ensure safer human-robot interaction due to their tiny form factor and weight -- i.e., tens of grams.
This compelling vision is challenged by the high level of intelligence needed aboard, which clashes against the limited computational and storage resources available on PULP (parallel-ultra-low-power) MCU class navigation and mission controllers that can be hosted aboard.
This work moves from PULP-Dronet, a State-of-the-Art convolutional neural network for autonomous navigation on nano-drones. 
We introduce Tiny-PULP-Dronet: a novel methodology to squeeze by more than one order of magnitude model size (50$\times$ fewer parameters),  and number of operations (27$\times$ less multiply-and-accumulate) required to run inference with similar flight performance as PULP-Dronet.
This massive reduction paves the way towards affordable multi-tasking on nano-drones, a fundamental requirement for achieving high-level intelligence.
\end{abstract}


\section{Introduction}\label{sec:introduction}

With their sub-10~\SI{}{\centi\meter} diameter and tens of grams in weight, nano-UAVs are agile and highly versatile robotic platforms, employed in many use cases where small size is crucial. 
From aerial inspection in narrow and dangerous places~\cite{uav_chemicals} to close-proximity human-robot interaction tasks~\cite{frontnet}, nano-UAVs are the ideal candidate platforms.
Similar to other autonomous robots, also nano-UAVs navigate thanks to the interaction of three key sub-systems~\cite{funflieber}. 
The \textit{state estimator} determines the current state of the system. 
Then, what we call the onboard \textit{intelligence} is responsible for solving the decision-making problem of choosing the next target state. 
Finally, the \textit{control} part brings the system from the current state to the target one.
Our work focuses on the intelligence, where the limited payload and power density of nano-drones clash with the onboard execution of complex real-time algorithms and even more with multi-tasking vision-based solutions.

Therefore, our goal is to minimize the onboard intelligence workload. 
This scenario allows to optimally balance between \textit{i)} enhancing the UAV's reactivity with an increased throughput in the decision-making process~\cite{framerate_in_cv_applications}, and \textit{ii)} freeing up resources for the execution of multi-perception intelligence tasks aboard our nano-drone.
These features enable nano-UAVs to tackle more complex flight missions.
For example, autonomous aerial cinematography and high-speed drone racing require a combination of multiple tasks running concurrently aboard~\cite{Bonatti_cinematic_uav, scaramuzza_alphapilot}, including visual-inertial odometry, object detection, trajectory planning, environment mapping, and collision avoidance.
Ultimately, we could significantly push small-size robotic platforms towards the biological levels of intelligence~\cite{bio_multitask_GIURFA1997505}. 

To date, nano-drones can accomplish individual intelligence autonomous navigation tasks by leveraging deep learning-based (DL) algorithms -- spanning a broad spectrum of complexity and sensorial input~\cite{inclined_landing, neural_swarm_2, frontnet, niculescu2021pulpdronetAICAS}.
In~\cite{inclined_landing}, Kooi and Babu\v{s}ka use a deep reinforcement learning approach with proximal policy optimization for the autonomous landing of a nano-drone on an inclined surface.
The designed convolutional neural network (CNN) requires about \SI{4.5}{\kilo\nothing} multiply-accumulate (MAC) operations per forward step, being sufficiently small to allow a single action evaluation in about \SI{2.5}{\milli \second} on a single-core Cortex-M4 processor, but still providing limited perception capabilities, restricted to landing purposes only.
Employing the same Cortex-M4, Neural-Swarm2~\cite{neural_swarm_2} exploits a DL-based controller to compensate close-proximity interaction forces that arise in formation flights of nano-drones.
With only about \SI{37}{\kilo MACs}, each nano-drone processes only the relative position and velocity of surrounding UAVs, enabling safe close-proximity flight.

Focusing on higher complexity DL-based workloads, the computational limitations imposed by single-core MCUs can be addressed by leveraging cutting-edge flight controllers targeting artificial intelligence.
An example of these new-generation devices is given by the 9-core GWT GAP8\footnote{https:// greenwaves-technologies.com/gap8\_mcu\_ai}, a parallel-ultra-low-power (PULP) processor.
This processor was previously exploited on UAVs in the PULP-Dronet project~\cite{niculescu2021pulpdronetAICAS}, leading to a fully-programmable end-to-end visual-based autonomous navigation engine for nano-drones.
PULP-Dronet is a single CNN capable of navigating a 27-grams nano-drone in both indoor and outdoor environments by predicting a collision probability, for obstacle avoidance, and a steering angle, to keep the drone within a lane.
Similarly, in the PULP-Frontnet project~\cite{frontnet} the PULP paradigm is leveraged to successfully run a lightweight CNN (down to \SI{4.8}{\mega MAC} and \SI{78}{\kilo\byte}) that performs a real-time relative pose estimation of a free-moving human, on a nano-UAV. 
This prediction is then fed to the nano-drone's controller, enabling precise ``human following'' capability.
Both PULP-Dronet and PULP-Frontnet demonstrate the feasibility of embedding high-level intelligence aboard nano-UAVs.
However, all these State-of-the-Art (SoA) solutions focus on vision-based single task intelligence, as the computational/memory burden for multi-tasking perception is still challenging ultra-low-power MCUs.

This paper presents a general methodology for analyzing the size and complexity of deep learning models suffering from \textit{overfitting} and \textit{sparsity}, which we apply, as an example, to the SoA PULP-Dronet CNN.
Then, we study the various trade-offs between the number of channels, pruning of inactive neurons, and accuracy, demonstrating the effectiveness of our method by introducing novel squeezed versions of the PULP-Dronet called Tiny-PULP-Dronets.
Our CNNs are up to 50$\times$ smaller and 8.5$\times$ faster than the baseline running on the same PULP GAP8 SoC, with no compromise on the final regression/classification performance.
Ultimately, the Tiny-PULP-Dronets, with a minimum model size of only \SI{6.4}{\kilo\byte} and a maximum throughput of more than \SI{160}{frame/\second}, enable higher reactivity on the autonomous navigation task, and leave sufficient memory and compute headroom for onboard multi-tasking intelligence even at the nano-sized scale.

\section{Methodology} \label{sec:methodology}

PULP-Dronet~\cite{niculescu2021pulpdronetAICAS} is a ResNet-based~\cite{resnet2015} CNN made of three residual blocks (ResBlocks), where each one consists of a main branch, performing two $3\times3$ convolutions, and a parallel by-pass (Byp), performing one $1\times1$ convolution.
This CNN produces a steering angle (regression) and a collision probability (classification) output.
Therefore, the model is trained using two different metrics: the mean squared error (MSE) and the binary cross-entropy (BCE).
The two metrics are then combined in a single loss function, $Loss = MSE + \beta BCE$, where $\beta$ is set to 0 for the first 10 epochs, gradually increasing in a logarithmic way to prioritize the regression problem.
Finally, the training process exploits a dynamic \textit{negative hard-mining} procedure, gradually narrowing down the loss computation to the k-top samples accounting for the highest error.

PULP-Dronet is deployed in fixed-point arithmetic on a multi-core GAP8 SoC, yielding \SI{41}{\mega MAC} operations per frame and \SI{320}{\kilo\byte} of weights.
The peak memory footprint -- also including input and intermediate buffers -- is as much as \SI{400}{\kilo\byte}, which is close to the total on-chip L2 memory (i.e., \SI{512}{\kilo\byte}) and represents a strong limiting factor for our multi-tasking objective.

\begin{figure}[tb]
 \includegraphics[width=1\linewidth]{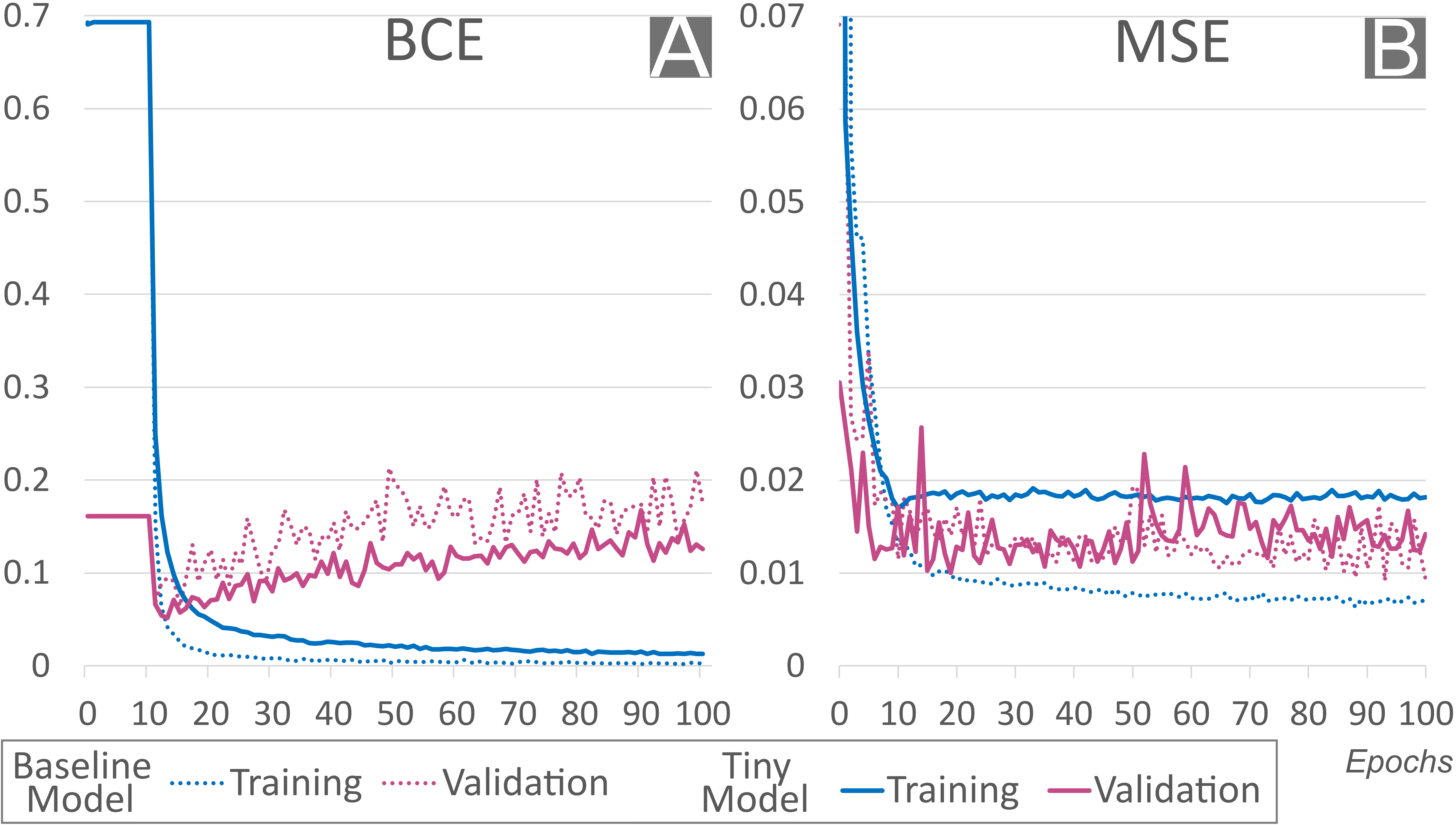}
 \caption{Training/validation loss' components -- BCE (A) and MSE (B) -- of PULP-Dronet, comparing the the baseline model against the tiny one.}
 \label{fig:overfitting}
\end{figure}

\textbf{Overfitting.}
A strong indicator of a deep learning model's overfitting is a decreasing/constant training loss paired with a validation loss that grows over the epochs.
In Figure~\ref{fig:overfitting}, we show both BCE (A) and MSE (B) curves over 100 epochs for both training (\SI{64}{\kilo\nothing} images) and validation (\SI{7}{\kilo\nothing} images) procedure.
In Figure~\ref{fig:overfitting}-A, the \textit{baseline} PULP-Dronet (dotted lines) suffers from overfitting on the BCE, showing a constantly growing validation error, while the MSE (Figure~\ref{fig:overfitting}-B) shows a noisy but almost constant validation error trend.
This behavior gives us the intuition that the network topology could be over-parameterized for the given perception problem.

Therefore, we introduce a model exploration to minimize its size while aiming at the same minimum in the validation losses -- or even lower.
Our approach aims at thinning the network's tensors by applying a $\gamma$ scaling factor to the number of channels across all layers.
We span the $\gamma$ parameter in the range $[0.125, 0.250, 0.5, 1.0]$, where $\gamma=1.0$ corresponds to the baseline PULP-Dronet, and we call the smaller variants Tiny-PULP-Dronet CNNs.
The smallest Tiny-PULP-Dronet ($\gamma=0.125$) losses are reported in Figure~\ref{fig:overfitting} (solid lines), where both the baseline and the tiny model assess to the same minimum validation MSE of $0.01$.
On the other hand, the tiny model achieves a minimum validation BCE of $0.052$, slightly lower than the corresponding value of the baseline ($0.064$), confirming overfitting of the baseline PULP-Dronet.

\textbf{Sparsity.}
Deep learning models' sparsity analysis is a key tool for identifying and selectively pruning parts of the models that are not contributing to the learning process~\cite{hoefler2021sparsity}.
We define as the \textit{structural activation's sparsity} the percentage of neurons in a convolutional layer that never activate over the entire validation dataset -- always equal to $0$.
This analysis for the PULP-Dronet baseline is reported in Table~\ref{table:sparsity}, which highlights significant sparsity across the whole network, peaking in the ResBlock3 ($92\%$ for the last by-pass).
These results suggest: \textit{i)} the network might suffer of over-parametrization, across all layers; and \textit{ii)} by-pass branches, usually exploited in very deep CNNs to avoid \textit{vanishing gradient} effects~\cite{resnet2015}, are underutilized in this shallow CNN.

Therefore, we analyze the gradients for each network layer after the by-pass removal: the convolution associated with Act1 shows the strongest gradient's attenuation.
On this layer, we record on the last 10 epochs (worst case) an average gradient magnitude of $0.055$ with a standard deviation of $0.16$, supporting the initial intuition of no vanishing gradients on the PULP-Dronet shallow CNN.
Furthermore, in Table~\ref{table:sparsity}, we also provide the same sparsity analysis for the smallest Tiny-PULP-Dronet ($\gamma=0.125$), with by-pass branches removed.
Compared to the baseline model, the tiny one scores a sparsity of almost $0\%$ across all layers, suggesting a more efficient usage of the neurons left after the two squeezing techniques.

\begin{table}
\scriptsize
\caption{Structural sparsity analysis of PULP-Dronet, comparing the baseline vs. the tiny model.}
\label{table:sparsity}
\centering
\setlength{\tabcolsep}{0.45em} 
\renewcommand{\arraystretch}{1.1} 
\begin{tabular}{|l|c|c|c|c|c|c|c|c|c|c|} 
\cline{2-11}
\multicolumn{1}{l|}{} & \textbf{Conv} & \multicolumn{3}{c|}{\textbf{ResBlock1}} & \multicolumn{3}{c|}{\textbf{ResBlock2}} & \multicolumn{3}{c|}{\textbf{ResBlock3}} \\ 
\hline
Topology & Act0 & Act1 & Act2 & Byp1 & Act3 & Act4 & Byp2 & Act5 & Act6 & Byp3 \\ 
\hline
Baseline & 28\% & 13\% & 6\%  & 6\% & 25\% & 10\% & 11\% & 49\% & 60\% & 92\% \\ 
Tiny & 0.15\% & 0\% & 0\% & — & 0\% & 0.07\% & —& 0\% & 0\% & — \\
\hline
\end{tabular}
\end{table}

\section{Results} \label{sec:results}

\subsection{Models evaluation} \label{sec:results_model_shrinking}

\begin{figure}[t]
 \includegraphics[width=1\linewidth]{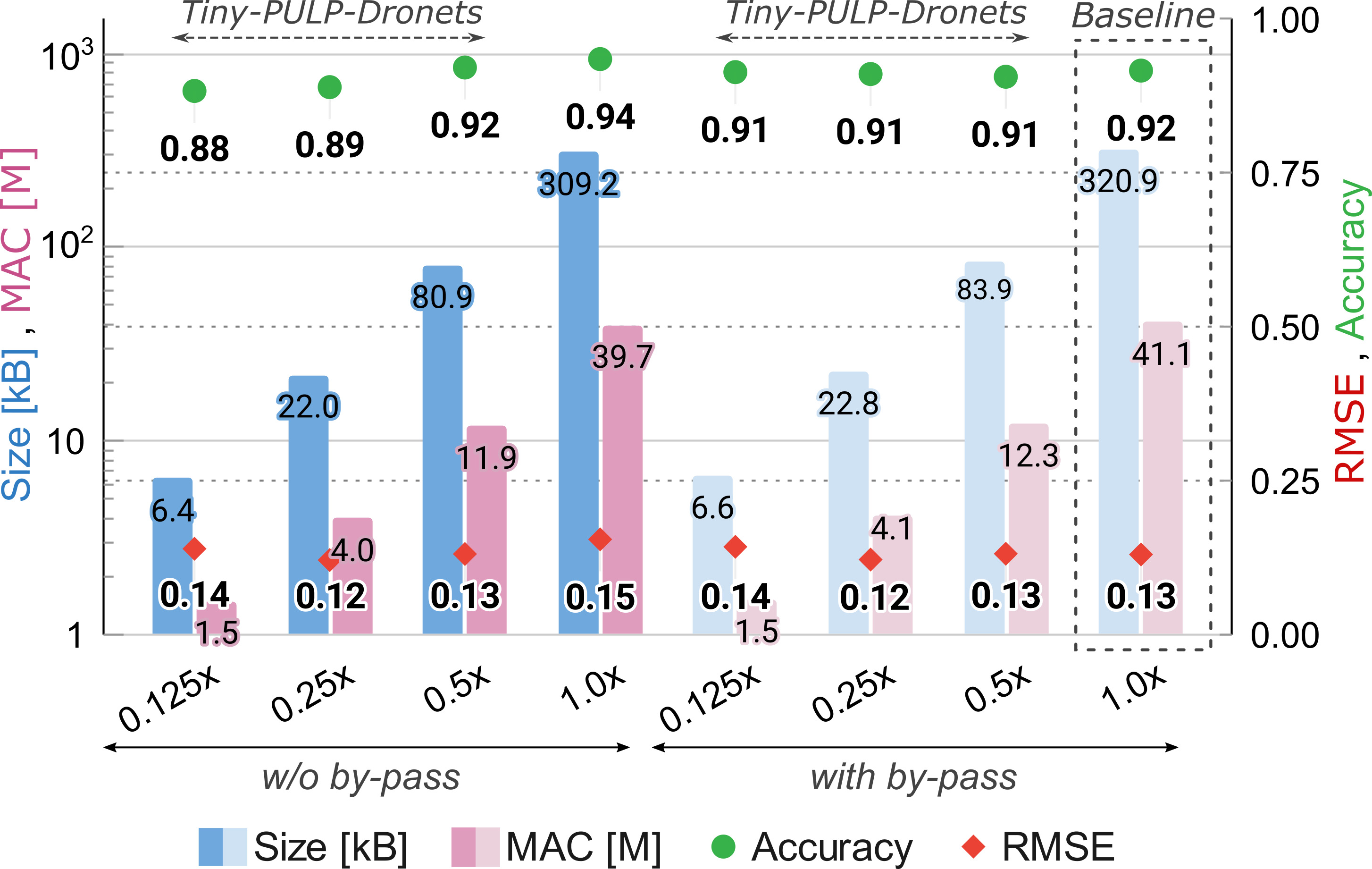}
 \caption{Comparing PULP-Dronet vs. its Tiny variants, in terms of size, MAC, Accuracy, and RMSE. We span $\gamma$ in the $[0.125, 0.250, 0.5, 1.0]$ range.}
 \label{fig:model_shrinking}
\end{figure}

In this section, we present our study on the effect of the proposed methodology when applied to the PULP-Dronet CNN in terms of memory footprint, computational effort, and regression/classification performance.
Figure~\ref{fig:model_shrinking} presents the comparison between the baseline CNN against its Tiny variants.
The reduction of the number of channels through $\gamma$ saves \numrange{237}{314}~\SI{}{\kilo\byte} and \numrange{29}{39}~\SI{}{\mega MAC}.
Additionally, also the by-pass removal reduces the memory usage of \numrange{1}{12}~\SI{}{\kilo\byte} and the operations by \numrange{0.1}{1}~\SI{}{\mega MAC}, depending on the specific model. 
Combining by-pass removal and the scaling of network's channels, the smallest Tiny-PULP-Dronet reduces $50\times$  memory footprint and $27\times$ on MAC operations vs. the baseline, reaching a minimum of \SI{6.4}{\kilo\byte} and \SI{1.5}{\mega MAC}, respectively.

Focusing on the regression performance for the testing set, the root mean squared error (RMSE) shows a similar trend for both \textit{with} and \textit{without} (w/o) by-pass groups (Figure~\ref{fig:model_shrinking}).
The score slowly improves while reducing the model size from $\gamma=1$ to $\gamma=0.25$, which is a common behavior of those models that suffer from overfitting and take advantage of an increased generalization capability, reducing their trainable parameters~\cite{hoefler2021sparsity}.
Instead, for both groups, the smallest models ($\gamma=0.125$) show a lower improvement on the RMSE, suggesting their potential underfitting.
Figure~\ref{fig:model_shrinking} also reports the Accuracy evaluation for the classification problem.
While this metric, on the \textit{with by-pass} group, seems to keep an almost constant performance all over the sizes ($\sim$0.91), it slowly reduces with the size for the \textit{without by-pass} group.
Overall, the by-pass removal does not significantly penalize either the RMSE or the Accuracy of the models, allowing to reach a remarkable 0.88 Accuracy for the smallest configuration. More importantly, this pruning is highly desirable for \textit{i}) reducing the memory footprint ($\sim3\%$) and operations ($\sim1.5\%$), and \textit{ii}) simplifying the deployment process on MCUs.

To support the second point, we further analyze the peak memory allocation needed for one single-image inference of the CNNs accounting for input/output intermediate buffers and network weights.
Looking at the peak memory allocation utilizing a simple \textit{incremental allocator}, which sums up the memory required by each layer, the baseline PULP-Dronet requires \SI{870}{\kilo\byte} compared to only \SI{105}{\kilo\byte} for the smallest Tiny-PULP-Dronet ($\gamma=0.125$).
On the other hand, by employing a \textit{dynamic allocator} which maximizes data reuse, the baseline PULP-Dronet peaks at \SI{400}{\kilo\byte}, while the smallest CNN peaks at \SI{80.1}{\kilo\byte}.
In this latest case, the removal of by-pass branches simplifies and reduces the memory burden eliminating the need for the simultaneous storage of two intermediate activations along parallel branches, which peaks at \SI{10}{\kilo\byte} in the smallest Tiny-PULP-Dronet.
Considering the dynamic allocator, we ultimately reduce by $5\times$ the peak memory usage required by the onboard intelligence, a key element to enable multi-tasking aboard constrained devices like the GAP8 SoC.

\begin{figure*}[tb]
\includegraphics[width=1\linewidth]{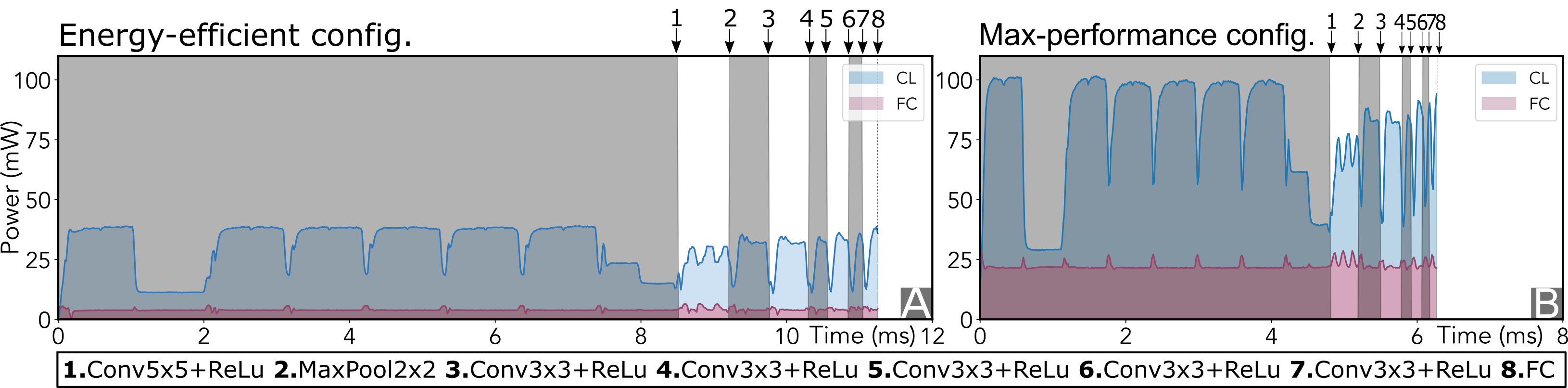}
    \caption{GAP8 power waveforms executing the smallest Tiny-PULP-Dronet ($\gamma=0.125$). (A) most energy-efficient SoC's configuration -- FC@\SI{50}{\mega \hertz}, CL@\SI{100}{\mega \hertz}, VDD@\SI{1.0}{\volt} -- and (B) the maximum performance one -- FC@\SI{250}{\mega \hertz}, CL@\SI{175}{\mega \hertz}, VDD@\SI{1.2}{\volt}.}
\label{fig:power_waveforms}
\end{figure*}

\subsection{Power analysis} \label{sec:power_analysis}

We evaluate GAP8's execution time and energy consumption when running a single-frame inference with the smallest Tiny-PULP-Dronet model ($\gamma=0.125$).
We use the RocketLogger data logger~\cite{sigrist2016rocketlogger} (\SI{64}{\kilo sps}) to separately plot the power waveforms of the GAP8's main core, the Fabric Controller (FC), and its 8-core parallel cluster (CL).
We test two SoC configurations: the so called energy-efficient configuration~\cite{niculescu2021pulpdronetAICAS}, whose operating point is FC@\SI{50}{\mega \hertz}, CL@\SI{100}{\mega \hertz}, and VDD@\SI{1}{\volt}, and the maximum performance settings according to the GAP8's datasheet, which is  FC@\SI{250}{\mega \hertz}, CL@\SI{175}{\mega \hertz}, VDD@\SI{1.2}{\volt}.

GAP8 takes \SI{1.1}{\mega cycles} to process the Tiny-PULP-Dronet, as shown with the power traces in Figure~\ref{fig:power_waveforms}, where we highlight the individual execution of each network's layer, for both SoC configurations.
The energy-efficient configuration (Figure~\ref{fig:power_waveforms}-A) shows an average power consumption of \SI{34}{\milli \watt}, \SI{11.3}{\milli \second} inference time, for a total energy consumption of \SI{0.38}{\milli \joule}. 
Moving to the maximum performance configuration (Figure~\ref{fig:power_waveforms}-B), the inference time gets reduced to \SI{6.3}{\milli\second} under the same average power, leading to \SI{0.63}{\milli \joule} per frame.
Overall, these results show an improvement of $8.5\times$ on the network's inference time when compared to the baseline PULP-Dronet, which takes \SI{96}{\milli \second} and \SI{52}{\milli \second} to run an inference under the energy-efficient and maximum performance SoC configurations, respectively.
Ultimately, we improve the frame-rate of PULP-Dronet from \SI{10}{frame/\second} to \SI{89}{frame/\second} on the energy efficient configuration, and from \SI{19}{frame/\second} to \SI{160}{frame/\second}, on the maximum performance one. 

Lastly, we analyze the execution of the first $5\times5$ convolution, which takes about $75\%$ of the total network's execution time on the smallest Tiny-PULP-Dronet ($\gamma=0.125$). 
This layer processes a $200\times200\times1$ input image and outputs a $100\times100\times4$ feature map, resulting in \SI{1}{\mega MAC} operations, which corresponds to the $70\%$ of the total network's operations, partially explaining its higher execution time over the others.
However, alongside the reduction by $27\times$ of the network's MAC operations w.r.t. the baseline, we only witness a latency reduction of $8.5\times$. 
This non-linear scaling is due to inevitable non-idealities: \textit{i}) the Height-Width-Channel data layout limits the input feature map data reuse, being proportional to the layer's output channels number, i.e., only 4, \textit{ii}) the marshaling stage required by the convolution for padding and constructing the input's flattened buffer adds up to $45\%$ of the total layer execution.
Ultimately, with a peak $8.5\times$ throughput improvement, Tiny-PULP-Dronets enable a faster reactivity of the nano-UAV when autonomously navigating the environment.

\section{Conclusion} \label{sec:conclusion}

The limited payload and computational power of nano-UAVs prevent high-level onboard intelligence, such as multi-tasking execution, which is still out of reach. 

This work presents a general methodology that leverages the CNN's sparsity and overfitting to squeeze both memory footprint and computational effort.
By applying our methodology to the SoA PULP-Dronet CNN for autonomous driving, we introduce its Tiny-PULP-Dronet variants.
These CNNs show a reduced memory burden ($50\times$ smaller) and computational complexity ($27\times$ lower) vs. the original model while \textit{i)} preserving the same regression performance, \textit{ii)} having minimal accuracy degradation ($6\%$ at most), and ultimately leaving sufficient resources for faster and lighter inference on multi-tasking autonomous nano-drones.

\section*{Acknowledgments}
We thank A. Giusti and A. Burrello for their support.

\bstctlcite{IEEEexample:BSTcontrol}

\bibliographystyle{IEEEtran}
\bibliography{IEEEabrv,bibliography}

\end{document}